\documentclass[conference]{IEEEtran}
\IEEEoverridecommandlockouts

\usepackage{cite}
\usepackage{amsmath,amssymb,amsfonts}
\usepackage{algorithmic}
\usepackage{graphicx}
\usepackage{textcomp}
\usepackage{xcolor}

\usepackage{inconsolata}
\usepackage{color}
\usepackage{multirow}
\usepackage{booktabs}
\usepackage{pifont}
\usepackage{svg}
\usepackage{listings}
\usepackage{float}
\usepackage[labelfont=bf]{caption}
\def\BibTeX{{\rm B\kern-.05em{\sc i\kern-.025em b}\kern-.08em
    T\kern-.1667em\lower.7ex\hbox{E}\kern-.125emX}}
\begin{document}

\title{Enhancing Large Language Models with Pseudo- and Multisource- Knowledge Graphs for Open-ended Question Answering}

\author{\IEEEauthorblockN{Jiaxiang Liu}
\IEEEauthorblockA{Institute of Automation, \\Chinese Academy of Sciences \\
University of Chinese Academy of Sciences\\
Beijing, China\\
liujiaxiang21@mails.ucas.ac.cn}
\and
\IEEEauthorblockN{Tong Zhou}
\IEEEauthorblockA{Institute of Automation, \\Chinese Academy of Sciences \\
Beijing, China\\
tong.zhou@ia.ac.cn}
\and
\IEEEauthorblockN{Yubo Chen}
\IEEEauthorblockA{Institute of Automation, \\Chinese Academy of Sciences \\
University of Chinese Academy of Sciences\\
Beijing, China\\
yubo.chen@nlpr.ia.ac.cn}
\and
\IEEEauthorblockN{Jun Zhao}
\IEEEauthorblockA{Institute of Automation, \\Chinese Academy of Sciences \\
University of Chinese Academy of Sciences\\
Beijing, China\\
jzhao@nlpr.ia.ac.cn}
\and
\IEEEauthorblockN{Kang Liu}
\IEEEauthorblockA{Institute of Automation, \\Chinese Academy of Sciences \\
University of Chinese Academy of Sciences\\
Beijing, China\\
kliu@nlpr.ia.ac.cn}
}

\maketitle

\begin{abstract}
Mitigating the hallucinations of Large Language Models is a crucial task. Although some existing methods employ self-enhancement techniques, they fall short of effectively addressing unknown factual hallucinations. Meanwhile, Knowledge Graph (KG) enhancement approaches fail to address the generalization across different KG sources and the enhancement of open-ended answer questions simultaneously. To tackle these limitations, we propose a framework that combines \textbf{P}seudo-Graph \textbf{G}eneration and \textbf{A}tomic \textbf{K}nowledge \textbf{V}erification (\textbf{PG\&AKV}). Enhancement of open-ended question-answering begins with leveraging the Pseudo-Graph Generation to provide the related knowledge framework. Subsequently, Atomic Knowledge Verification utilizes atomic-level knowledge querying and verification to achieve generalizability under different KG sources. Compared to the baseline, this approach yields a minimum improvement of 11.5 in the ROUGE-L score for open-ended questions. For precise-answered questions, we observe a minimum accuracy improvement of 7.5\%. Moreover, PG\&AKV also exhibits generalizability across different KG sources. Utilizing KG different from the question sources, PG\&AKV can even achieve at least a 3.5 \% performance improvement. In summary, our results pave the way for enhancing LLMs by incorporating Pseudo- and Multisource-KGs, particularly in the filed of open-ended questions.
\end{abstract}

\begin{IEEEkeywords}
Retrieval-augmented generation, knowledge-enhanced LLMs, open-ended QA.
\end{IEEEkeywords}

\section{Introduction}
\begin{table*}
\caption{A comparison of several enhanced methods is outlined as follows: 1) Train-free, meaning no additional training steps are required; 2) QID-free and Relation-free, meaning the method does not rely on entities' QID or corresponding relations; 3) Knowledge-enhanced, meaning the method uses external knowledge to improve an LLM; 4) Multi-source, meaning it generalizes well across various KG sources; 5) Robustness, meaning early errors minimally impact later steps; 6) Open-ended QA, meaning it can enhance LLMs for questions with open-ended answers.
}
\centering \scalebox{1}{
\begin{tabular}{c c c c c c c c}\toprule[2pt] \hline
			\multirow{2}{*}{Methods}& \multicolumn{6}{c}{Comparison dimension} \\ \cmidrule(lr){2-8}
            &Train-free&QID-free&Relation-free&Knowledge-enhanced&Multi-cource& Robustness&Open-ended QA  \\ \hline
            CoT&\textcolor{green}{\ding{52}}&
            \textcolor{green}{\ding{52}}&
            \textcolor{green}{\ding{52}}&
            \textcolor{red}{\ding{56}}&
            \textcolor{red}{\ding{56}}&
            \textcolor{red}{\ding{56}}&
            \textcolor{green}{\ding{52}} \\ 
            RAG&\textcolor{green}{\ding{52}}&
            \textcolor{green}{\ding{52}}&
            \textcolor{green}{\ding{52}}&
            \textcolor{green}{\ding{52}}&
            \textcolor{red}{\ding{56}}&
            \textcolor{green}{\ding{52}}&
            \textcolor{green}{\ding{52}}\\ 
            SQL-PALM&\textcolor{red}{\ding{56}}&
            \textcolor{red}{\ding{56}}&
            \textcolor{green}{\ding{52}}&
            \textcolor{green}{\ding{52}}&
            \textcolor{red}{\ding{56}}&
            \textcolor{red}{\ding{56}}&
            \textcolor{red}{\ding{56}}\\ 
            ToG&\textcolor{green}{\ding{52}}&
            \textcolor{red}{\ding{56}}&
            \textcolor{red}{\ding{56}}&
            \textcolor{green}{\ding{52}}&
            \textcolor{green}{\ding{52}}&
            \textcolor{red}{\ding{56}}&
            \textcolor{red}{\ding{56}}\\ 
            KGR&\textcolor{green}{\ding{52}}&
            \textcolor{green}{\ding{52}}&
            \textcolor{red}{\ding{56}}&
            \textcolor{green}{\ding{52}}&
            \textcolor{red}{\ding{56}}&
            \textcolor{green}{\ding{52}}&
            \textcolor{red}{\ding{56}}\\ 
            \textbf{Ours}&\textcolor{green}{\ding{52}}&
            \textcolor{green}{\ding{52}}&
            \textcolor{green}{\ding{52}}&
            \textcolor{green}{\ding{52}}&
            \textcolor{green}{\ding{52}}&
            \textcolor{green}{\ding{52}}&
            \textcolor{green}{\ding{52}}\\ \hline
            
            \bottomrule[2pt]
		\end{tabular}}
        \label{related-work}
\end{table*}

Large language models (LLMs) \cite{GPT3, GPT3-5, GPT4, touvron2023llama2, chowdhery2022palm} have achieved remarkable results in the field of question answering tasks. They obtain the capability to handle various questions through a large scale of pre-training data. However, LLMs still face issues of hallucination and lack of specific domain knowledge when dealing with complex problems \cite{huang2023survey, ye2023cognitive}. 

To mitigate the hallucination of models and thus improve the accuracy of models' responses, various methods have been proposed. 

Some approaches leverage the model's own capabilities to address uncertain knowledge. Chain-of-Thought (CoT) prompting \cite{CoT} method encourages the model to generate intermediate reasoning steps, thereby enhancing the accuracy. Self-Consistency (SC) \cite{Self-Consistency} method enhances the robustness of CoT by considering a synthesis of model's multiple reasoning processes. Nevertheless, these techniques do not fundamentally resolve the hallucination issue of LLMs. Because such issues often arise from errors or gaps in the training data \cite{ye2023cognitive}. As a result, incorporating external knowledge sources has emerged as a promising strategy to further mitigate hallucinations in LLMs.

The second approach is to use knowledge graphs (KGs) to enhance LLMs. KGs, like Wikidata\footnote{www.wikidata.org}, DBpedia\footnote{https://www.dbpedia.org/} and YAGO\footnote{https://yago-knowledge.org/}, are highly valued in LLMs tasks due to their structured knowledge, high accuracy, and timely updates of information \cite{pan2024unifying}. Therefore, how to extract knowledge from KGs to enhance large models is an important researched field. A straightforward approach is to prompt \cite{chang2023prompt} or fine tune \cite{sun2023sql} LLMs to generate Structured Query Language (SQL). However, the schema of different knowledge graphs are different, limiting the generalization ability of this method. To address the challenge of generalization across different KGs, one strategy involves semantically encoding the graphs and enhancing LLMs with retrieval-based methods \cite{NEURIPS2020_6b493230}. However, this approach may struggle with open-ended questions where key retrieval elements are not explicitly provided. For instance, when asked “Who is the most famous painter in the world?”, RAG method \cite{NEURIPS2020_6b493230} may only find data points such as $<X> <occupation> <painter>$ in Wikidata. This information alone is insufficient to answer related questions because it lacks significant details, like the artist’s notable works. Another approach is to enhance generalization by utilizing LLMs, such as KGR \cite{KGR} and ToG \cite{ToG}. Although methods like KGR and ToG have achieved promising results, each has its limitations. ToG leaks the QID of entities in the KG, while KGR requires specific relations for the corresponding entities. These issues compromise the models' ability to generalize effectively in real-world applications such as open-ended question answering, where the QID and the relations are not given explicitly. In summary, the previous KG-enhanced LLM methods cannot simultaneously address the following two problems: 1) generalization across different knowledge graphs, and 2) utilizing KG to enhance open-ended question answering.

To address the above issues, we propose a two-phase approach called \textbf{P}seudo-Graph \textbf{G}eneration and \textbf{A}tomic \textbf{K}nowledge \textbf{V}erification (\textbf{PG\&AKV}). In Pseudo-Graph Generation, PG\&AKV leverages LLMs to generate pseudo-triples relevant to the question. This step enables LLMs to use their general knowledge to identify the information needed to answer open-ended questions. Even if LLMs occasionally produce hallucinations during this process, they still effectively construct the framework of the necessary knowledge, guiding subsequent queries. For Atomic Knowledge Verification, we use the generated pseudo-triples to perform semantic querying on vectorized KGs. Because both the querying and verification operate at an atomic level independent of any specific KG schema, this approach generalizes well across different KGs. Finally, LLMs are utilized to verify the pseudo-triples based on the queried triples, resulting in the desired answer. Overall, PG\&AKV mitigates factual hallucinations by enhancing LLMs outputs with reliable external knowledge from diverse KGs.

\begin{figure*}
		\centering
		\includegraphics[width=1.0\textwidth]{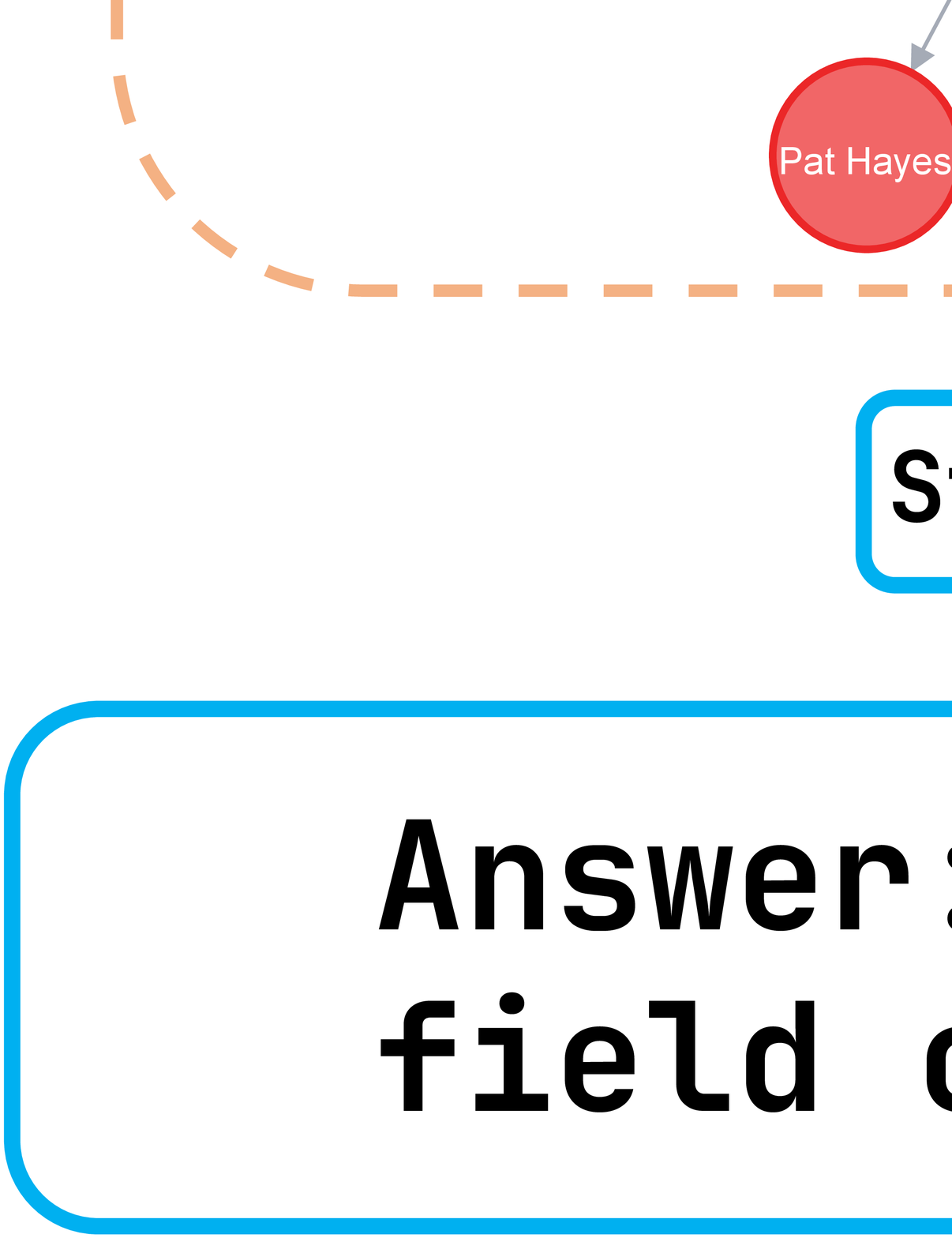}
		\caption{\label{pipline}
  The over view of PG\&AKV: \textbf{Pseudo-Graph Generation:} In step 1, we prompt LLM to generate pseudo-graph $G_p$ related to the question. \textbf{Atomic Knowledge Verification:} For step 2, the pseudo-triples extracted from $G_p$ are used to query a semantic KG, yielding the ground truth graph ($G_g$). In step 3, the LLM verifies $G_p$, which leads to the fixed graph ($G_f$).
  }
\end{figure*}

Our contributions are listed below:
\begin{itemize}
\item[$\bullet$] The Pseudo-Graph Generation leverages the LLMs to generate pseudo-triples relevant to the question, making it possible to utilize KG as knowledge augmentation for LLMs in open-ended question answering. 

\item[$\bullet$] Atomic Knowledge Verification uses atomic-level knowledge. Therefore, it ensures generalisability of PG\&AKV across different KGs

\item[$\bullet$] We introduce an open-ended question-answering dataset in a KG-enhanced setting named Nature Questions. Our experimental results show that our method not only performs excellently on this dataset but also demonstrates strong performance on existing precise-answered datasets such as QALD-10\cite{qald} and SimpleQuestions\cite{SQA}. 
\end{itemize}

\section{Related Work}
Table~\ref{related-work} demonstrates a comparison of the capabilities of the existing methods.
\subsection{Self Enhanced LLMs}
Directly fine-tuning LLMs to improve performance is challenging due to the enormous computational resources required. Chain-of-Thought (CoT) \cite{CoT} prompt method was shown to stimulate factual knowledge in LLMs by generating explicit reasoning steps during generation process. This innovation has sparked a series of follow-up studies. For instance, Zero-shot-CoT \cite{kojima2022large} uses the prompt "Let’s think step by step" to elicit effective reasoning, while Auto-CoT \cite{zhang2023automatic} automates the construction of high-quality reasoning sequences. The Self-Consistency (SC) \cite{Self-Consistency} method considers a synthesis of multiple models' reasoning processes. Additionally, approaches like Knowledge-driven CoT \cite{wang2023knowledge} and KAM-CoT \cite{mondal2024kam} integrate external knowledge into the CoT framework.

\subsection{KG Enhanced LLMs}
Simply enhancing LLMs through prompts is far from sufficient. For example, when questions require new or updated knowledge not present in the model's training data, simple prompting falls short. In contrast, leveraging KGs—which offer structured information, high accuracy, and timely updates \cite{pan2024unifying}—provides a practical way to bolster LLM performance.

Existing work focuses on how to accurately extract knowledge from KGs. A straightforward approach is to prompt \cite{chang2023prompt} or fine tune \cite{sun2023sql} LLMs to generate Structured Query Language (SQL). However, the prompt-based method faces challenges because LLMs struggle to generate correct SQL without being provided with specific entity QID. For instance, when asked, “Please tell me what is the QID of Yellow River in Wikidata?”, ChatGPT returns Q1826, despite the correct QID being Q2066882. On the other hand, the fine-tuning approach not only requires significant computational resources but also often suffers from limited generalizability due to the varying schemas of different KGs. 

Embedding representations like TransE \cite{TransE} offer a promising alternative for handling these schema differences. However, this approach may struggle with open-ended questions
where key information are not explicitly provided \cite{li-etal-2023-copy}. 

ToG \cite{ToG} and KGR \cite{KGR} are proposed to leverage LLMs for identifying key information points. ToG utilizes the model to search for relevant relations based on the entities to solve complex problems. While, KGR identifies relevant entities from the answers of LLMs. However, these methods all exhibit ambiguity in the retrieval steps. ToG directly leaks the entities' QID, while KGR requires the corresponding relations. This significantly weakens the generalizability of these methods in practical applications, where the
QID and relations are not provided. To the best of our knowledge, methods that enhance LLMs using KGs have not yet been applied to open-ended QA.

\section{Methodology}
In this section, we would like to describe the process of PG\&AKV. The general flow of the method can be seen in Fig. \ref{pipline}. First, we define a triple of the form $G = \{S, R, O\}$, where $S$ denotes the set of subjects, $R$ is the set of relations, and $O$ is the set of objects.
\subsection{Pseudo-Graph Generation}
\begin{figure}[!ht]
		\centering
		\includegraphics[width=0.45\textwidth]{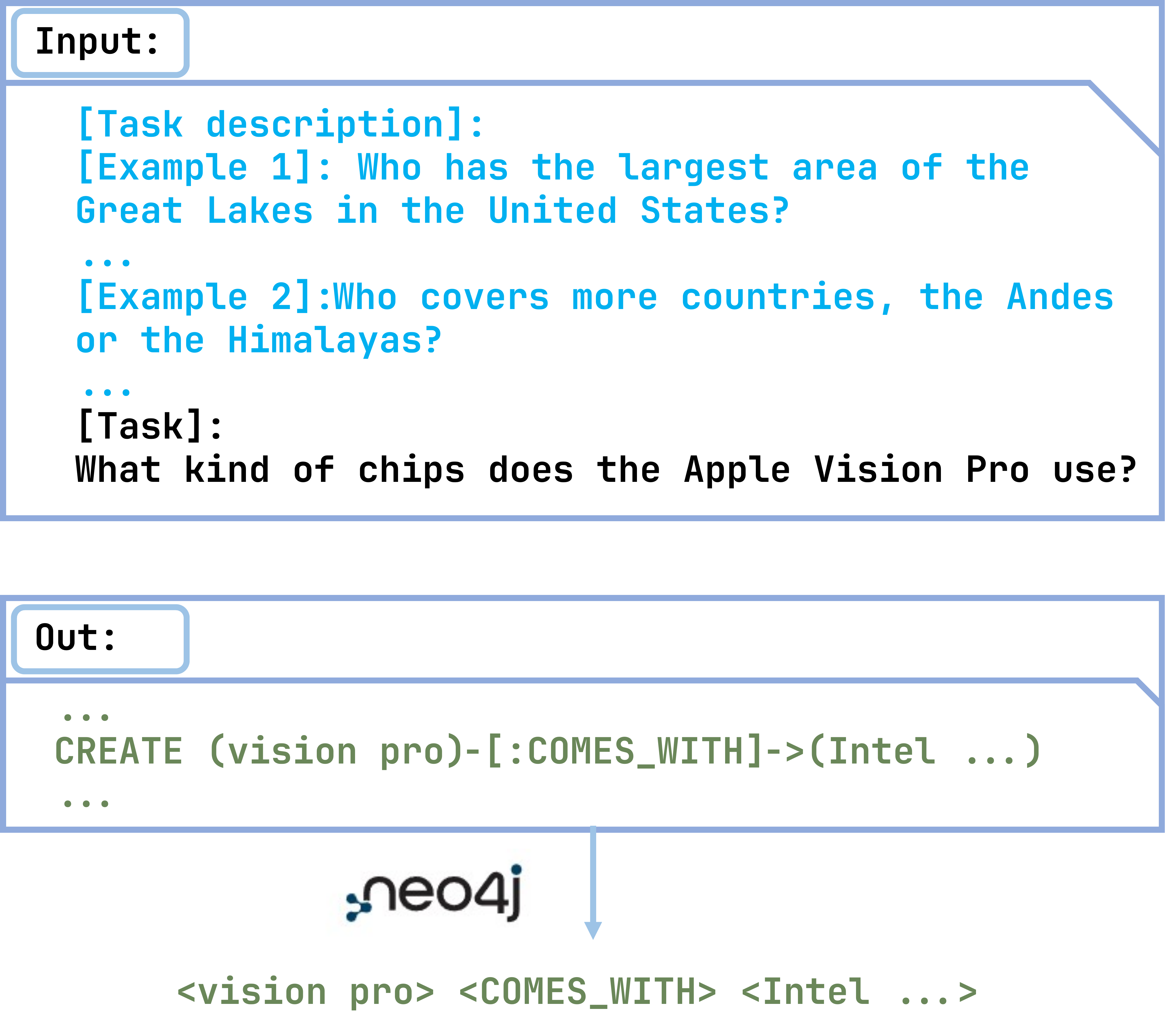}
		\caption{Pipline of the generation of pseudo-graph.}
        \label{neo4j}
\end{figure}\par
For processing of Pseudo-Graph Generation, our initial approach was to directly generate fact-related triples. However, without fine-tuning, it is challenging for LLMs to accurately produce structural triples. Since LLMs are trained on large-scale natural corpora \cite{GPT3, GPT4, touvron2023llama2, chowdhery2022palm}, they naturally tend to generate continuous language rather than the discrete, rule-bound format triples. This may result in outputs that do not conform to the necessary structure, causing errors in subsequent steps ($<Allen Newell> <made\quad Sora>$). Considering that LLMs also display strong code-generation capabilities \cite{yetiştiren2023evaluating}, PG\&AKV uses programming languages as an intermediary bridge between unstructured natural language and structured triple.

As Fig. \ref{neo4j} shows, firstly, the LLM is provided with two examples in the form of Cypher language, which instruct it to generate Cypher queries designed to answer the question. Next, we execute these queries on Neo4j\footnote{www.neo4j.com} and decode the results into pseudo-graph $G_p = \{S_p, R_p, O_p\}$. This approach enables the model to generate pseudo-graphs with an accuracy of 98\%, while the accuracy of directly generation is only 75\%.

\subsection{Atomic Knowledge Verification}
\paragraph{Sematic Query} We generate a semantic knowledge graph by extracting subsets from Wikidata or Freebase. Next, Sentence-BERT \cite{reimers-gurevych-2019-sentence} is used to encode the semantic triples into $G_{base}$ for retrieval.

We then calculate the cosine similarity between each triple in graph $G_p$ and every triple in $G_{base}$. For each triple in $G_p$, we select the top 10 most similar triples from $G_{base}$, forming a temporary graph $G_t = \{S_t, R_t, O_t\}$. However, $G_p$ may contain a large number of triples, the size of $G_t$ can be extensive or even exceeding the maximum token limit of the LLM. Therefore, a pruning method is necessary.

We propose a two-step pruning method that overcomes the limitations of relying solely on LLM judgments, as in the ToG \cite{ToG} method. And it efficiently leverages the $G_p$ produced by the LLM. 

\textbf{Step 1-Candidate Selection:} Rather than using LLM-based scoring to prune relations, our method begins by selecting the top k entities from the set $S_t$, where k matches the size of $S_p$ in $G_p$. Specifically, we rank entities according to the number of triples in which they appear and choose the highest-ranked candidates. This step effectively filters out entities that provide less information for the question.

\textbf{Step 2-Semantic Ranking:} After the initial filtering, we use cosine similarity in the semantic query phase to further refine and rank the candidates. For each subject $s \in S_t$, we calculate an \textbf{\textit{entity confidence score}} by averaging the cosine similarity scores of all triples in $G_t$ that use $s$ as the subject. Entities receiving a confidence score below 0.7 are then filtered out.

The result of these two steps is the ground truth graph $G_g$, which accurately represents the relevant information for answering the query without accumulating errors from overreliance on LLM judgments.

\paragraph{Pseudo-Graph Verification}

As shown in Fig. \ref{Verification}, when verifying $G_p$ with LLM, the subjects of entity $s \in S_g$ with higher \textbf{\textit{entity confidence score}} are place closer to $G_p$ in the context. This approach is more advantageous for LLM to establish 
better attention between the $G_p$ and $G_g$. In our prompt, there are two simple examples used to enable LLM to perform self-verification of the knowledge graph. Afterwards, we obtain the fixed knowledge graph $G_f$.

\subsection{Answer Generation}
In this step, we use two examples to teach LLM how to answer the questions based on the knowledge graph. Then, LLM is instructed to generate answers to questions using the question and $G_f$. Our prompt is shown in Fig. \ref{generation}.

\section{Experimens}
Our experiments mainly aim to answer the following two questions: i) How does the performance of PG\&AKV compare to other methods; ii) The generalization across different KG sources of PG\&AKV.
\begin{table*}
\caption{
The main results of our experiments. \textbf{Bold} indicates the best-performing method on the dataset for the model. SimpleQuestions and QALD use Hit@1 for evaulation. And Nature Questions evals with ROUGE-L. Since the data structure of ToG is not suitable for Nature Questions, we do not list the result of ToG on Nature Questions.
}
\centering \scalebox{1}{
\begin{tabular}{c c c c c}\toprule[2pt] \hline
\multicolumn{2}{c}{\multirow{2}*{Method}}&\multicolumn{2}{c}{Hit@1}&ROUGE L \\ 
\cmidrule(lr){3-4} \cmidrule(lr){5-5}
&&SimpleQuestions&QALD-10&Nature Questions \\ \hline
\multirow{5}*{GPT-3.5}
&ToG&45.4&48.6&-\\ \cmidrule(lr){2-5}
&IO&20.2&38.7&20.5 \\
&CoT&22.0&40.5&23.2\\
&SC&21.2&41.1&23.5 \\
&RAG&27.5&34.2&23.8 \\
&Ours&\textbf{34.3}&\textbf{48.6}&\textbf{37.5} \\ \hline
\multirow{5}*{GPT-4}
&ToG&58.6&54.7&-\\ \cmidrule(lr){2-5}
&IO&29.9&44.7&20.9 \\
&CoT&32.2&48.9&27.7 \\
&SC&36.0&48.9&27.6 \\
&RAG&31.3&46.2&27.0 \\
&Ours&\textbf{40.0}&\textbf{56.5}&\textbf{39.2} \\ \hline

            \bottomrule[2pt]
		\end{tabular}}
        \label{result}
\end{table*}

\subsection{Models}
In our experiments, we used GPT-3.5 and GPT-4 \cite{GPT4} to generate pseudo-graph, perform atomic knowledge verification, and produce the final answers. Sentence-BERT \cite{plenz-etal-2023-similarity} is chosen as the encoder for the semantic KG and as the query module to query knowledge related to the triples generated by the LLM.

\subsection{Datasets}
To verify the effectiveness of PG\&AKV, our experiment is conducted on three different types of datasets, including single-hop questions, multi-hop questions, and open-answer questions: SimpleQuestions \cite{SQA}, QALD-10 \cite{qald}, Nature Questions.

\begin{itemize}
\item[$\bullet$] SimpleQuestions \cite{SQA} employs a manually annotated method to generate corresponding questions based on the facts in the Freebase.

\item[$\bullet$]  QALD-10 \cite{qald} is a multilingual, multi-hop question answering dataset that uses Wikidata as its knowledge base. In our experiments, English is choose for the question-and-answer tasks.

\item[$\bullet$] Nature Questions is a dataset we compiled, featuring questions in daily life that include open-ended answers, multiple-answer responses, and queries about new knowledge. We manually constructed 50 questions for this dataset, writing three answers for each question, expecting the answer will be comprehensive enough.
\end{itemize} 

\textbf{Detail} Since the Freebase API has been closed, we use a subset of FB2M \cite{bordes2015largescale} to serve as our Freebase-KG. Because the SimpleQuestions dataset \cite{SQA} contains a large number of data points (100k), we randomly selected a subset for testing. For both QALD-10 \cite{qald} and Nature Questions, we use the full dataset for testing and constructing the corresponding semantic KG based on the questions.

\textbf{Evaluation Metrics} Regarding evaluation metrics, for both SimpleQuestions \cite{SQA} and QALD-10 \cite{qald}, we adopted the Hit@1 metric as the measure of question answering accuracy. For the Nature Questions dataset, ROUGE-L-f1 \cite{lin-2004-rouge} is used to evaluate the accuracy and comprehensiveness of LLM's answers.

\subsection{Baselines}
In order to judge the validity of the Pseudo-Graph Generation and Atomic Knowledge Verification method, the following baselines for comparison are chosen:
\begin{itemize}
\item[$\bullet$] \textbf{IO} \cite{GPT3}: We use the standard input-output (IO) prompt as for conducting direct testing of the model, with 6 in-context examples.

\item[$\bullet$] \textbf{Chain of Thougnt} (CoT) \cite{CoT} It encourages the model to generate the reasoning process, with 6 in-context examples.

\item[$\bullet$]  \textbf{Self-Consistency} (SC) \cite{Self-Consistency}: We use a sampling temperature of 0.7 and perform three sampling iterations, using voting to process the results in our experiments. 

\item[$\bullet$]  \textbf{Retrieval-Augmented Generation} (RAG) \cite{NEURIPS2020_6b493230}: We directly matching the question with the semantic KG for retrieval.

\item[$\bullet$]  \textbf{Think on Graph} (ToG) \cite{ToG}: Although ToG does not match our experimental setup, we still list the results for comparison.
\end{itemize}

\subsection{Main Results}
Our main results can be seen in Table~\ref{result}. It demonstrates the effectiveness of the framework in open-ended questions and in traditional precise questions.
\subsubsection{Comparison With Other Methods}
Firstly, we can observe from Table~\ref{result} that the PG\&AKV method achieves better results than the baselines across different LLMs and datasets. In contrast, RAG performs the worst on QALD-10 with GPT-3.5, achieving results that are 4.5\% lower than those of the IO method. This gap suggests that directly matching the semantics of a question to knowledge triples, especially in multi-hop questions, is challenging.

\textbf{Advantage in deterministic question answering:} For deterministic question answering, PG\&AKV method achieves the highest accuracy on the QALD-10 dataset. On the QALD-10 dataset, PG\&AKV even outperforms the ToG method, which achieves a result of 54.7\% with GPT-4\cite{ToG}. Furthermore, PG\&AKV also achieves a higher accuracy than the fine-tuned SOTA\cite{Borroto2022SPARQLQAET}, whose accuracy is 45.4\%. The improvement is also evident on the SimpleQuestions, where PG\&AKV also achieves improvements of 10.9\% (GPT-3.5) and 5.3\% (GPT-4) compared to the second-best baseline on the two models. We also found that, with the enhancement of the approach, the factual hallucination of GPT-3.5 can be effectively addressed. It even outperforms various GPT-4 baselines on SimpleQuestion dataset.

\textbf{Better performance in Nature Questions answering:} On the Nature Questions answering dataset, PG\&AKV also shows significant improvement in the ROUGE-L evaluation metric. With the help of PG\&AKV, for open-ended problems, GPT-3.5 performs even better than GPT-4-CoT. Moreover, GPT-4 achieves an improvement of 11.5 points in ROUGE-L. These findings suggest that our method is effective for real-world applications. Taking into account the improvement from QALD-10 (multi-hop questions) and Nature Questions (open-ended questions), we conclude that PG\&AKV is particularly effective for tackling complex questions.

In conclusion, the above results indicate that good performance in both precise question-answering and open-ended question-answering tasks has been achieved by LLMs with PG\&AKV.

\subsubsection{Generalization Across Different KG Sources}
We demonstrate the generalization capability of PG\&AKV across different KG sources by evaluating it on the same set of questions using varied KG sources. For this test, we chose GPT-3.5 as the model and validated the method using both the SimpleQuestions and Nature Questions datasets. We then compared the model’s performance improvements over the CoT baseline.

Table~\ref{diff-KGs} clearly shows that PG\&AKV outperforms the CoT method on the same set of questions across various KG sources. It is important to note that while the SimpleQuestions dataset uses Freebase as its knowledge graph, some relations that are single-hop in Freebase require multi-hop reasoning in Wikidata. This discrepancy was not accounted for during the construction of the knowledge graph, leading to a smaller performance improvement. We leave this for subsequent investigation.

\begin{table}[!ht]
    \caption{Performance on SimpleQuestions and Nature Questions with different KG sources. The SimpleQuestions is based on the Freebase as KG sources.}
    \centering\scalebox{1}{
    \begin{tabular}{ccc}
    \toprule[2pt]\hline 
        \multirow{2}*{\textbf{Method}}&\multirow{2}*{\textbf{SimpleQuestion}} &\multirow{2}*{\textbf{Nature Questions}} \\ 
        && \\ \hline
        CoT&22.0 &23.2 \\ \hline
        Our/ Freebase&38.2 &26.7 \\ 
        Gain&\textcolor{green}{+16.2} &\textcolor{green}{+3.5} \\ \hline
        Our/ Wikidata&28.1 &37.5 \\ 
        Gain&\textcolor{green}{+6.1} &\textcolor{green}{+14.3} \\ \hline
        \bottomrule[2pt]
    \end{tabular}  
    }
    \label{diff-KGs}
\end{table}

In conclusion, the results above have demonstrated the generalization capability of PG\&AKV across different KG sources.

\subsection{Ablation Study}
In this section, we will explore the functionality of each component of PG\&AKV. Our findings are: i) Pseudo-Graph Generation stimulate model's knowledge; ii) Atomic Knowledge Verification increases precision and breadth of knowledge. To investigate this, we selected GPT-3.5 and GPT-4 for testing on the QALD-10 and Nature Questions. During the experiment, we compared the model's accuracy when directly providing the pseudo-graph $G_p$ against using the fixed graph $G_f$.
\begin{table}[!ht]
    \caption{GPT3.5's performance on QALD-10 and Nature Questions with different references.}
    \centering\scalebox{1}{
    \begin{tabular}{ccc}
    \toprule[2pt]\hline 
        \multirow{2}*{\textbf{Method}}&\multirow{2}*{\textbf{QALD-10}} &\multirow{2}*{\textbf{Nature Questions}} \\ 
        && \\ \hline
        CoT&40.5&23.2 \\ \hline
        w/ $G_p$&44.4&24.3 \\
        Gain from CoT&\textcolor{green}{+3.9}&\textcolor{green}{+1.1} \\ \hline
        w/ $G_h$&48.6&37.5 \\
        Gain from CoT&\textcolor{green}{+8.1}&\textcolor{green}{+14.3} \\ \hline
        
        \hline 
    \bottomrule[2pt]
    \end{tabular}  
    }
    \label{study-GPT3.5}
\end{table}

\begin{table}[!ht]
    \caption{GPT4's performance on QALD-10 and Nature Questions with different references.}
    \centering\scalebox{1}{
    \begin{tabular}{ccc}
    \toprule[2pt]\hline 
        \multirow{2}*{\textbf{Method}}&\multirow{2}*{\textbf{QALD-10}} &\multirow{2}*{\textbf{Nature Questions}} \\ 
        && \\ \hline
        CoT&48.9&27.7 \\ \hline 
        Pseudo-Graph&53.9&24.4 \\
        Gain from CoT&\textcolor{green}{+5.0}&\textcolor{red}{-3.3} \\ \hline
        Verification&56.5&39.2 \\
        Gain from CoT&\textcolor{green}{+7.6}&\textcolor{green}{+11.5} \\ \hline
        
        \hline 
        \bottomrule[2pt]
    \end{tabular}  
    }
    \label{study-GPT4}
\end{table}
\textbf{Pseudo-Graph Generation stimulate model's knowledge:} From Table~\ref{study-GPT3.5} , for GPT-3.5, generating pseudo-graphs rather than using the traditional CoT approach appears to better activate the model's factual knowledge. Similarly, Table \ref{study-GPT4} shows that integrating pseudo-graphs leads to improved performance on the QALD-10 dataset.

\textbf{Atomic Knowledge Verification increases precision and breadth of knowledge:} Additionally, from Table \ref{study-GPT3.5}, the verification steps did not reduce the model's performance; in fact, they improved the accuracy of the pseudo-graph. However, Table \ref{study-GPT4} shows that for Nature Questions, the pseudo-graph approach slightly decreased performance with GPT-4. This decline may occur because, when generating the pseudo-graph, the model tends to include only the information it is most certain about, resulting in a less comprehensive presentation of facts. The minor drop observed with the pseudo-graph further highlights the effectiveness of the Atomic Knowledge Verification steps. Additionally, we find that the main errors in the model's verification process were caused by LLM directly appending the base graph after the pseudo-graph and not making modifications to the pseudo-graph.

These results can to some extent explain the function of the two steps. Utilizing the model to generate pseudo-graphs can more effectively stimulate the model's self-retrieval capability compared to the traditional CoT method. Additionally, the Atomic Knowledge Verification step can not only correct erroneous facts in precise questioning but also significantly enhance the factual accuracy and comprehensiveness of the model's answers in open-ended questions.

\section{Conclusion \& Future Work}
We propose a framework that combines Pseudo-Graph Generation with Atomic Knowledge Verification to enhance LLMs in addressing open-ended questions. Pseudo-Graph Generation leverages the feature of hallucinate-even under erroneous conditions it can also provides us with a framework for knowledge points. In parallel, Atomic Knowledge Verification performs atomic-level semantic querying and verification, mitigating generalization issues across different KGs.  Experimental results indicate that our approach not only performs well on traditional precision-based queries but also significantly improves natural question-answering. Meanwhile, we demonstrate the functionality of two modules through ablation experiments: Pseudo-Graph Generation stimulate model's knowledge and Atomic Knowledge Verification increases precision and variegation of knowledge. Our approach points out a feasible direction for the enhancement of LLM using KG in practical applications.

For future research, we plan to improve semantic querying by incorporating more advanced embedding models. Additionally, we aim to explore alternative pruning strategies to enhance the quality of acquired knowledge. We are also considering the development of a dedicated Pseudo-Graph Verification module to further enrich the language model’s understanding. Finally, our goal is to engineer a practical framework that effectively augments large language models in real-world applications.


\bibliographystyle{IEEEtran}
\bibliography{custom}
\vspace{12pt}
\section{Appendix}
\subsection{Our prompts}
\begin{figure*}
		\centering
		\includegraphics[width=1\textwidth]{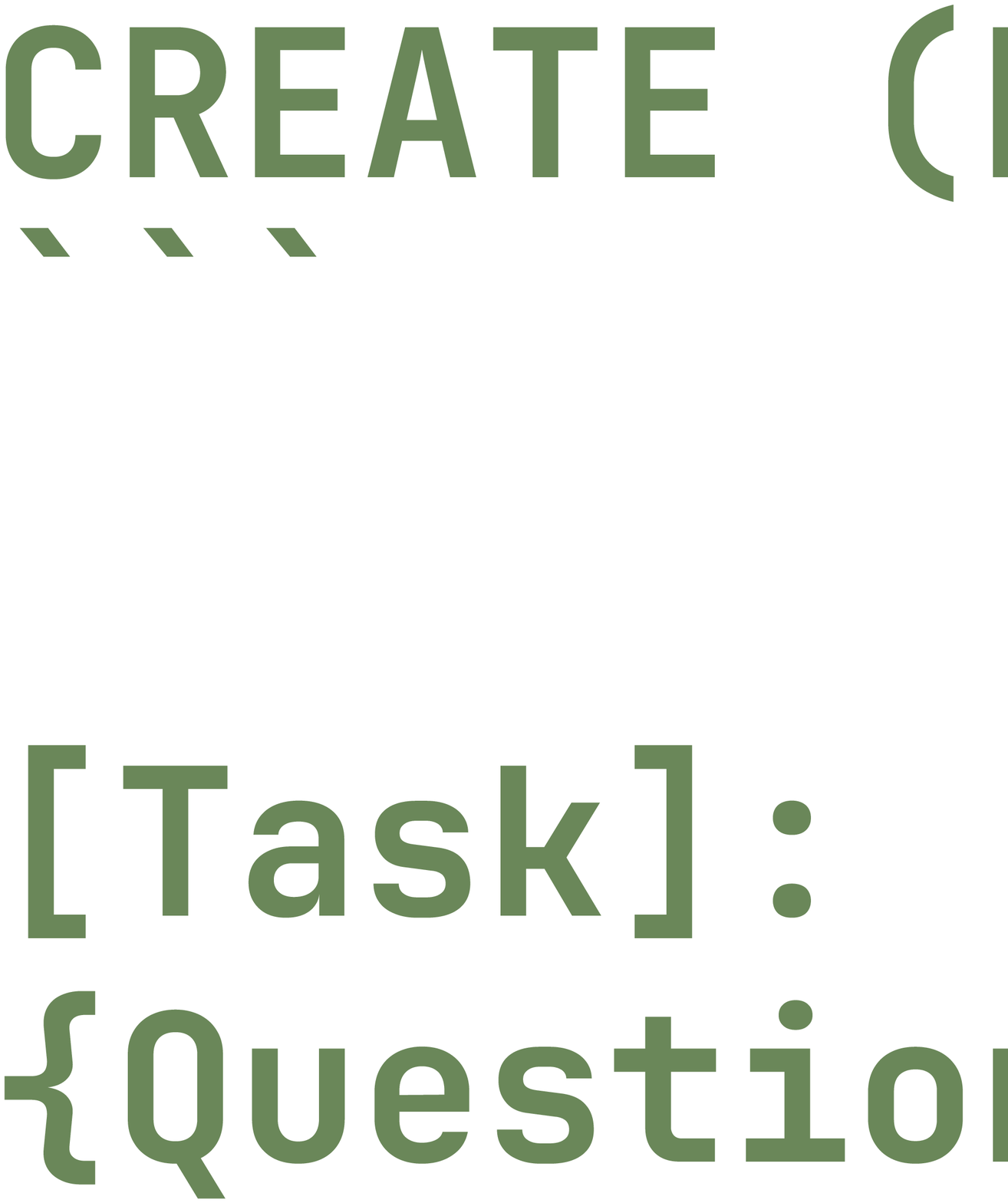}
		\caption{\label{p1}
        Prompt for pseudo-graph generation. We partially omit the section involving generated code due to the large number of lines it occupies.
  }
\end{figure*}

\begin{figure*}
		\centering
		\includegraphics[width=1\textwidth]{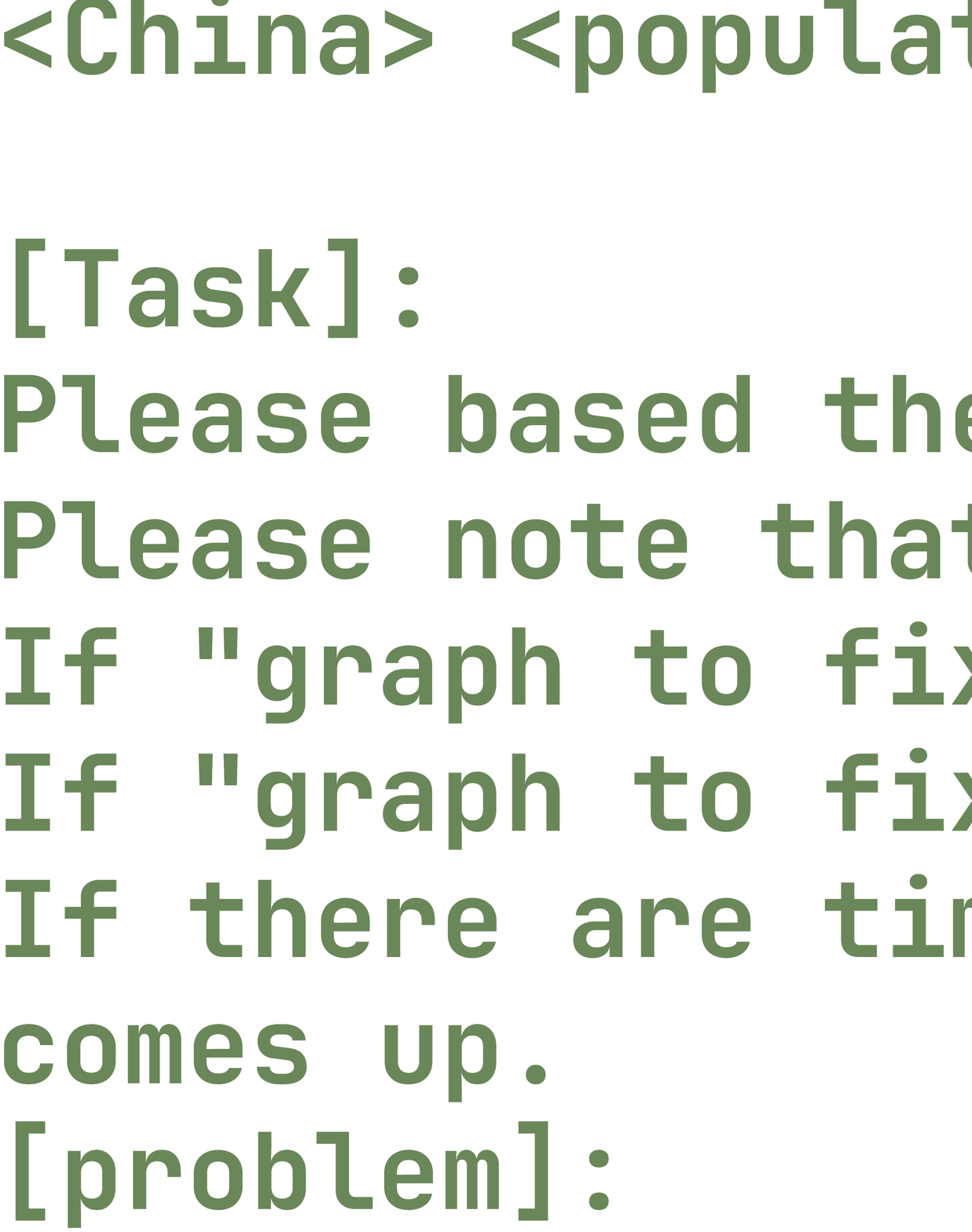}
		\caption{\label{Verification}Prompt for Pseudo-Graph Verification.}
\end{figure*}

\begin{figure*}
		\centering
		\includegraphics[width=1\textwidth]{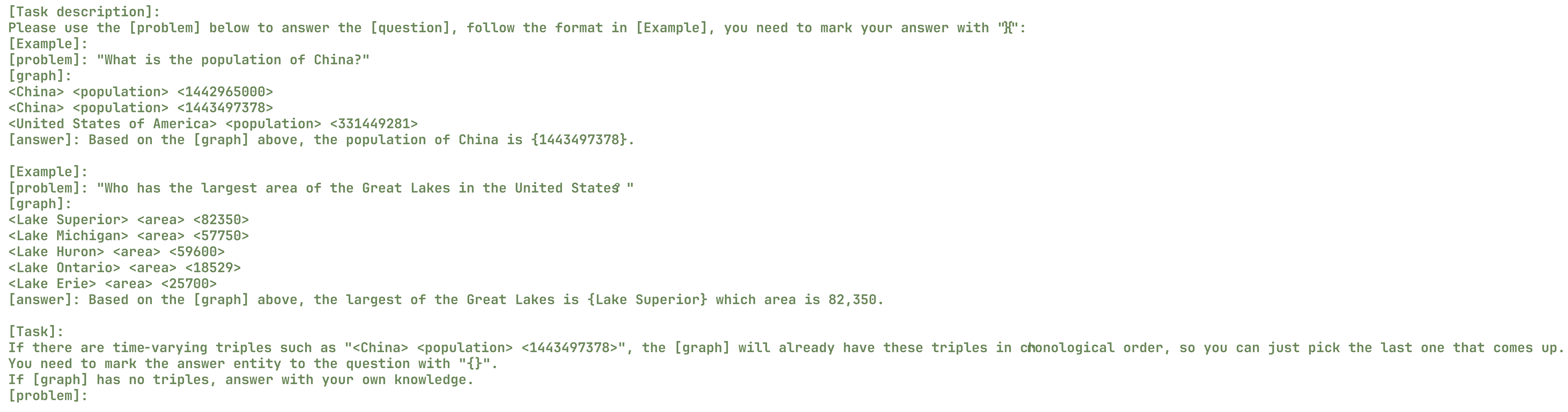}
        
		\caption{\label{generation}
        Prompt for answer generation.
  }
\end{figure*}

Our prompts can be seen in the Figure~\ref{p1},~\ref{Verification},~\ref{generation}.

\end{document}